\documentclass[10pt,twocolumn,letterpaper]{article}

\usepackage{cvpr}              

\usepackage{graphicx}
\usepackage{amsmath}
\usepackage{amssymb}
\usepackage{booktabs}

\usepackage{epsfig}
\usepackage{graphicx}
\usepackage{amsmath}
\usepackage{amssymb}
\usepackage{multirow}
\usepackage{enumerate}
\usepackage{bm}
\usepackage{wrapfig}
\usepackage{epstopdf}
\usepackage{wrapfig}

\usepackage[pagebackref,breaklinks,colorlinks]{hyperref}

\usepackage[capitalize]{cleveref}
\crefname{section}{Sec.}{Secs.}
\Crefname{section}{Section}{Sections}
\Crefname{table}{eTable}{Tables}
\crefname{table}{Tab.}{Tabs.}

\def\cvprPaperID{2757} 
\def\confName{CVPR}
\def\confYear{2022}

\begin{document}

\title{Reconstructing Surfaces for Sparse Point Clouds with On-Surface Priors}


\author{Baorui Ma, Yu-Shen Liu\thanks{The corresponding author is Yu-Shen Liu. This work was supported by National Key R$\&$D Program of China (2018YFB0505400, 2020YFF0304100), the National Natural Science Foundation of China (62072268), and in part by Tsinghua-Kuaishou Institute of Future Media Data.}\\
School of Software, BNRist, Tsinghua University\\
Beijing, China\\
{\tt\small mbr18@mails.tsinghua.edu.cn, \tt\small liuyushen@tsinghua.edu.cn}
\and
Zhizhong Han\\
Wayne State University\\
Detroit, USA\\
{\tt\small h312h@wayne.edu}
}
\maketitle

\begin{abstract}
It is an important task to reconstruct surfaces from 3D point clouds. Current methods are able to reconstruct surfaces by learning Signed Distance Functions (SDFs) from single point clouds without ground truth signed distances or point normals. However, they require the point clouds to be dense, which dramatically limits their performance in real applications. To resolve this issue, we propose to reconstruct highly accurate surfaces from sparse point clouds with an on-surface prior. We train a neural network to learn SDFs via projecting queries onto the surface represented by the sparse point cloud. Our key idea is to infer signed distances by pushing both the query projections to be on the surface and the projection distance to be the minimum. To achieve this, we train a neural network to capture the on-surface prior to determine whether a point is on a sparse point cloud or not, and then leverage it as a differentiable function to learn SDFs from unseen sparse point cloud. Our method can learn SDFs from a single sparse point cloud without ground truth signed distances or point normals. Our numerical evaluation under widely used benchmarks demonstrates that our method achieves state-of-the-art reconstruction accuracy, especially for sparse point clouds. Code and data are available at
\href{https://github.com/mabaorui/OnSurfacePrior}{https://github.com/mabaorui/OnSurfacePrior}.
\end{abstract}


\section{Introduction}
Reconstructing surfaces from 3D point clouds is a vital task in 3D computer vision. It bridges the gap between the data capturing and the surface editing for various downstream applications. It has been studied for decades using geometric approaches~\cite{journals/tog/KazhdanH13,Lorensen87marchingcubes,817351TVCG,OhtakeBATS03}. However, these methods require extensive human interaction to set proper parameters for different 3D point clouds, which leads to poor generalization ability. Therefore, the data-driven strategy becomes more promising to resolve this problem.

Recent learning based methods~\cite{jiang2020lig,Zhizhong2021icml,Peng2020ECCV,ErlerEtAl:Points2Surf:ECCV:2020,Liu2021MLS} leverage this strategy to learn signed distance functions (SDFs) from 3D point clouds, and further leverage the learned SDFs to reconstruct surfaces using the marching cubes algorithm~\cite{Lorensen87marchingcubes}. One kind of these methods~\cite{jiang2020lig,ErlerEtAl:Points2Surf:ECCV:2020,Liu2021MLS} requires supervision including ground truth signed distances or point normals during training, and infers SDFs for unseen 3D point clouds during test. To remove the requirement of the ground truth supervision, another kind of methods~\cite{gropp2020implicit,Atzmon_2020_CVPR,Zhizhong2021icml} can directly learn SDFs from single unseen 3D point cloud with geometric constraints~\cite{gropp2020implicit,Atzmon_2020_CVPR} or neural pulling~\cite{Zhizhong2021icml}. One key factor that makes these methods successful without the ground truth supervision is that the single point cloud should be dense, which supports to estimate the zero level set~\cite{gropp2020implicit} or search accurate pulling targets~\cite{Zhizhong2021icml}. However, due to the high cost of dense point clouds capturing, the assumption of dense point clouds fails in real applications. Therefore, it is appealing but challenging to learn SDFs from sparse point clouds without ground truth signed distances or normals.

To resolve this issue, we introduce to learn SDFs from single sparse point clouds with an on-surface prior. For a surface represented by a sparse point cloud, we aim to perceive its surrounding signed distance field via projecting an arbitrary query location onto the surface. Our novelty lies in the two constraints that we add on the projections, so that each projection locates on the surface and is the nearest to the query. This leads to two losses to train a neural network to learn SDFs. One loss is provided by the on-surface prior, which determines whether a projection is on the surface represented by the sparse point cloud or not, even if the projection is not a point of the sparse point cloud. While the other loss encourages the projection distance is the minimum to the surface. To achieve this, we train a neural network using a data-driven strategy to capture the on-surface prior from a dataset during training, and leverage the trained network as a differentiable function to learn SDFs for unseen sparse point clouds. For the learning of SDFs, our method does not require ground truth signed distances or point normals, and enables highly accurate surface reconstruction from sparse point clouds. We show our superior performance over the state-of-the-art methods by numerical and visual comparison under the widely used benchmarks. Our contributions are listed below.

\begin{enumerate}[i)]
\item We propose a method to learn SDFs from sparse point clouds without ground truth signed distances or point normals.
\item We introduce an on-surface prior which can determine the relationship between a point and a sparse point cloud, and further be used to train another network to learn SDFs.
\item Our method significantly outperforms the state-of-the-art methods in terms of surface reconstruction accuracy under large-scale benchmarks.
\end{enumerate}

\section{Related Work}
Deep learning-based 3D shape understanding has achieved very promising results in different tasks~\cite{Zhu2021NICESLAM,ruckert2021adop,Park_2019_CVPR,MeschederNetworks,mildenhall2020nerf,Zhizhong2018seq,Zhizhong2019seq,3D2SeqViews19,wenxin_2020_CVPR,seqxy2seqzeccv2020paper,Zhizhong2018VIP,Zhizhong2020icml,Han2019ShapeCaptionerGCacmmm,zhizhongiccv2021finepoints,wenxincvpr2022,MAPVAE19,p2seq18,hutaoaaai2020,wenxin_2021a_CVPR,wenxin_2021b_CVPR,Jiang2019SDFDiffDRcvpr,zhizhongiccv2021completing,tianyangcvpr2022,jain2021dreamfields,text2mesh,yu_and_fridovichkeil2021plenoxels,mueller2022instant}. Reconstructing surfaces from 3D point clouds is a classic research topic. Geometry based methods~\cite{journals/tog/KazhdanH13,Lorensen87marchingcubes,817351TVCG,OhtakeBATS03} tried to resolve this problem by analysing the geometry on the shape itself without learning experience from large scale dataset.

Recent learning based methods~\cite{ErlerEtAl:Points2Surf:ECCV:2020,Liu2021MLS,Peng2020ECCV,jia2020learning,jiang2020lig,DBLP:conf/eccv/ChabraLISSLN20,9320319Lombardi,Mi_2020_CVPR} achieve state-of-the-art results by learning various priors from dataset using deep learning models. Implicit functions, such as SDFs or occupancy fields, are usually learned to represent 3D shapes or scenes, and then the marching cubes algorithm is used to reconstruct the learned implicit functions into surfaces. Some methods~\cite{ErlerEtAl:Points2Surf:ECCV:2020,Liu2021MLS} require ground truth signed distances or point normals to learn global prior during training. Some other methods~\cite{Peng2020ECCV,jia2020learning,DBLP:journals/corr/abs-2105-02788,takikawa2021nglod} learn occupancy fields as a global prior using the ground truth occupancy supervision. To reveal more detailed geometry, local shape priors are learned as SDFs~\cite{jiang2020lig,DBLP:conf/eccv/ChabraLISSLN20,9320319Lombardi,Tretschk2020PatchNets} or occupancy fields~\cite{Mi_2020_CVPR} with supervision, where point clouds are usually split into different grids~\cite{jiang2020lig} or patches~\cite{Tretschk2020PatchNets} as local regions. Moreover, some more interesting methods for surface reconstruction are also proposed, such as a differentiable formulation of poisson solver~\cite{nipspoisson21}, retrieving parts~\cite{siddiqui2021retrievalfuse}, iso-points~\cite{yifan2021iso}, implicit moving least-squares surfaces~\cite{Liu2021MLS} or point convolution~\cite{pococvpr2022}.

Using meshing strategy, surfaces can also be reconstructed by connecting neighboring points using intrinsic-extrinsic metrics~\cite{liu2020meshing}, Delaunay triangulation of point clouds~\cite{luo2021deepdt} or connection from an initial meshes~\cite{Hanocka2020p2m}. Local regions represented as point clouds can also be reconstructed via fitting using the Wasserstein distance as a measure of approximation~\cite{Williams_2019_CVPR}.

More appealing solutions are to learn SDFs without ground truth signed distances or point normals. Some methods were proposed to achieve this using geometric constraints~\cite{gropp2020implicit,Atzmon_2020_CVPR,zhao2020signagnostic,atzmon2020sald,DBLP:journals/corr/abs-2106-10811,tang2021sign,DBLP:journals/corr/abs-2108-08931,sitzmann2019siren} or through neural pulling~\cite{Zhizhong2021icml}. However, these methods are limited by the assumption that point clouds are dense points, which makes them not perform well in real applications. Our method falls in this category, but differently, we can learn accurate SDFs from sparse point clouds. A cocurrent work~\cite{Needle3DPoints} resolves this problem without supervision.

\section{Method}
\noindent\textbf{Problem Statement. }Given a sparse point cloud $\bm{G}\in\mathbb R^{G\times 3}$, we aim to reconstruct its surface. We achieve this by learning SDFs $f_{\bm{\theta}}$ from $\bm{G}$ without requiring ground truth signed distances and normals of points on $\bm{G}$. $f_{\bm{\theta}}$ predicts signed distances $s=f_{\bm{\theta}}(\bm{q},\bm{c})$ for arbitrary queries $\bm{q}$ sampled around $\bm{G}$, where $\bm{c}$ is a condition identifying $\bm{G}$. Since our method can learn SDFs $f_{\bm{\theta}}$ from single point clouds, we will ignore the condition $\bm{c}$ in the following. With the learned $f_{\bm{\theta}}$, we reconstruct the surface of $\bm{G}$ using the marching cubes algorithm~\cite{Lorensen87marchingcubes}.

\begin{figure*}[tb]
  \centering
   \includegraphics[width=\linewidth]{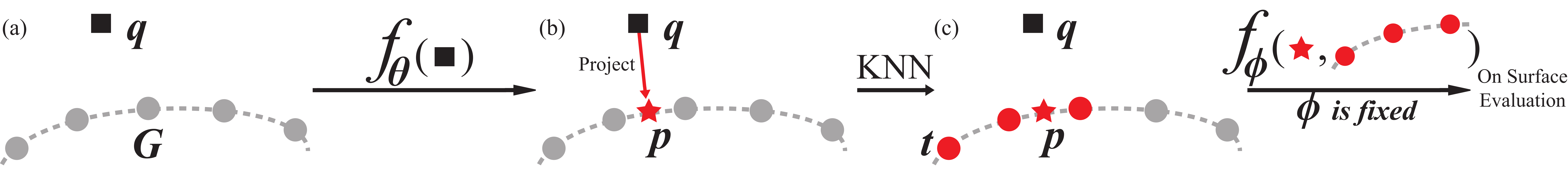}
  %
  %
  \vspace{-0.3in}
\caption{\label{fig:overview}The demonstration of our method during test. We leverage a data-driven strategy to learn on-surface decision function $f_{\bm{\phi}}$ as an on-surface prior during training. During testing, we learn SDFs $f_{\bm{\theta}}$ from a sparse point cloud $\bm{G}$. (a) Query $\bm{q}$ is sampled around $\bm{G}$. (b) $\bm{q}$ is projected into a projection $\bm{p}$ towards $\bm{G}$ using the path determined by SDFs $f_{\bm{\theta}}$. $f_{\bm{\phi}}$ evaluates whether $\bm{p}$ is on the surface represented by the $K$ nearest neighbors $\bm{t}$ of $\bm{p}$ in (c). We run the marching cubes to reconstruct surfaces of $f_{\bm{\theta}}$.}
\vspace{-0.15in}
\end{figure*}

\noindent\textbf{Overview. }Our method is demonstrated in a 2D case in Fig.~\ref{fig:overview}. It is mainly formed by two functions, i.e., a SDF $f_{\bm{\theta}}$ and an on-surface decision function (ODF) $f_{\bm{\phi}}$, both of which are learned by deep neural networks parameterized by $\bm{\theta}$ and $\bm{\phi}$, respectively. The SDF $f_{\bm{\theta}}$ learns the signed distance field around $\bm{G}$, with the on-surface prior provided by the ODF $f_{\bm{\phi}}$. Therefore, the parameters $\bm{\theta}$ in $f_{\bm{\theta}}$ are learned for single sparse point clouds with fixed parameters $\bm{\phi}$ in $f_{\bm{\phi}}$ during test, without the ground truth signed distances or point normals, while we learn $\bm{\phi}$ separately during training using a data-driven strategy.

We start from a query $\bm{q}$ around the sparse point cloud $\bm{G}$ in Fig.~\ref{fig:overview} (a). We project $\bm{q}$ towards $\bm{G}$ into a projection $\bm{p}$ in Fig.~\ref{fig:overview} (b), using the path determined by the signed distance $s$ and the gradient at $\bm{q}$ from the SDF $f_{\bm{\theta}}$. Then, we establish a local region $\bm{t}$ on $\bm{G}$ which is formed by the $K$ nearest neighbors of the projection $\bm{p}$ in Fig.~\ref{fig:overview} (c). Finally, the ODF $f_{\bm{\phi}}$ will determine whether the projection $\bm{p}$ is on the region $\bm{t}$ or not.

To learn $f_{\bm{\theta}}$, we penalize $f_{\bm{\theta}}$ through the differentiable function $f_{\bm{\phi}}$, if $f_{\bm{\phi}}$ determines that $\bm{p}$ is not on the region $\bm{t}$, and meanwhile, encourage $f_{\bm{\theta}}$ to produce the shortest path for the projection.

\noindent\textbf{Query Projection. }We project queries $\bm{q}$ onto sparse point cloud $\bm{G}$ as an evaluation of $f_{\bm{\theta}}$. If the projection path provided by $f_{\bm{\theta}}$ is correct, there would be no on-surface penalty on $f_{\bm{\theta}}$, and vice versa. The projection path for query $\bm{q}$ can be formed by the signed distance $s=f_{\bm{\theta}}(\bm{q})$ and the gradient $\nabla f_{\bm{\theta}}(\bm{q})$. This is also similar to the pulling procedure used in NeuralPull (NP)~\cite{Zhizhong2021icml}. The reason is that the absolute value of $s$ determines the distance from query $\bm{q}$ to the surface, and the normalized gradient $\bm{d}=\nabla f_{\bm{\theta}}(\bm{q})/||\nabla f_{\bm{\theta}}(\bm{q})||_2$ indicates the direction. Therefore, we can leverage the following equation to project query $\bm{q}$ to its projection $\bm{p}$ onto the surface that the zero-level set of $f_{\bm{\theta}}$ indicates,\vspace{-0.15in}

\begin{equation}
\label{eq:1}
\begin{aligned}
\bm{p}=\bm{q}-s\bm{d}.
\end{aligned}
\end{equation}

\begin{figure}[tb]
  \centering
   \includegraphics[width=\linewidth]{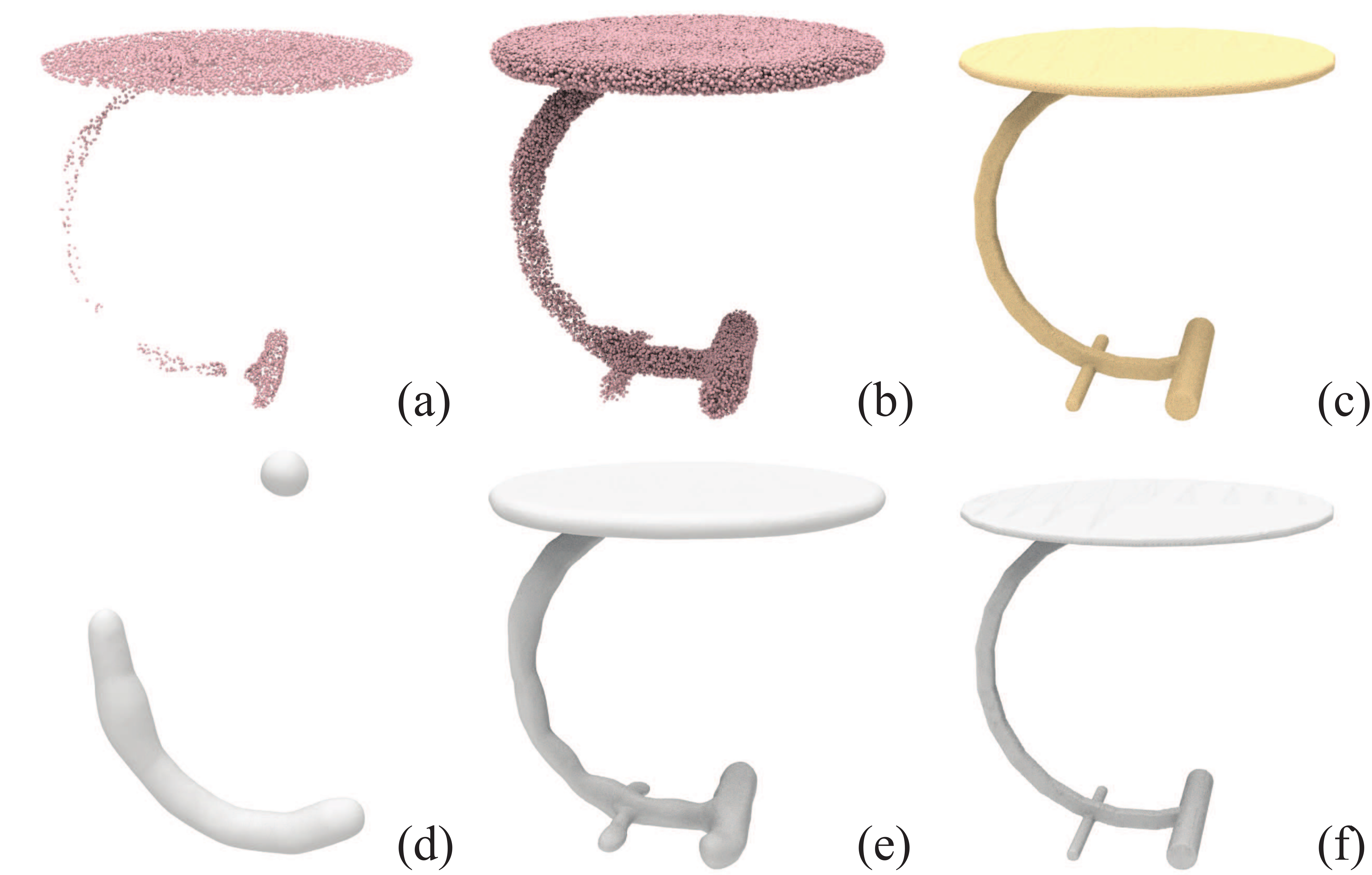}
  %
  %
  \vspace{-0.3in}
\caption{\label{fig:Surfacepoint}The visual comparison with binary classifier in (a) and unsigned distance function in (b) as the on-surface prior, and the visual comparison with (e) and without (d) geometric regularization. The ground truth shape and sampled points are shown in (f) and (c), respectively.}
\vspace{-0.3in}
\end{figure}

\noindent\textbf{On-Surface Prior. }The first constraint that we add on the projection $\bm{p}$ is that $\bm{p}$ should be located on the surface represented by the sparse point cloud $\bm{G}$. This is a difficult problem and the sparseness makes this problem even more difficult. An intuitive solution is first to establish a local patch $\bm{t}$ neighboring to $\bm{p}$, and fit a quadric surface on $\bm{t}$ like MPU~\cite{OhtakeBATS03}, and finally calculate the distance between $\bm{p}$ and the quadric surface to make a decision. However, fitting a quadric surface to a point cloud is still challenging, due to the sensitivity to parameter setting and geometry complexity, especially on sparse point clouds.

To resolve this issue, we leverage a data-driven strategy to learn a ODF $f_{\bm{\phi}}(\bm{p},\bm{t})$ using a deep neural network to determine whether the projection $\bm{p}$ is located on the surface of $\bm{t}$ or not. We expect $f_{\bm{\phi}}(\bm{p},\bm{t})$ to be class-agnostic and object-agnostic, so we regard $\bm{t}$ as a local patch rather than a global shape.

We first tried to learn $f_{\bm{\phi}}(\bm{p},\bm{t})$ as a binary classifier. The output of $f_{\bm{\phi}}$ indicates the probability of being on the surface. We prepare a training set $T=\{\{\bm{p}_i,\bm{t}_i,l_i\},i\in[1,I]\}$ with ground truth labels $l_i$ indicating whether a point $\bm{p}_i$ is on a specific point cloud region $\bm{t}_i$ or not. We leverage the public available benchmarks to obtain the dataset $T$. We sample $\bm{p}_i$ around each shape, and sample sparse points on the shape. Among the sampled sparse points, we regard the $K$ nearest neighbors of $\bm{p}_i$ as $\bm{t}_i$. We also label $\bm{p}_i$ to indicate whether it is sampled from the same surface as the region $\bm{t}_i$ or not.

However, our preliminary results show that it is very hard to learn a good $f_{\bm{\phi}}$. As demonstrated in Fig.~\ref{fig:Surfacepoint}, we learn $f_{\bm{\phi}}$ using one set of points $\bm{p}$ and evaluate it using another set of points $\bm{p}$, where both sets are sampled around the same shape in Fig.~\ref{fig:Surfacepoint} (c). We show points that $f_{\bm{\phi}}$ correctly classifies as on-surface points in Fig.~\ref{fig:Surfacepoint} (a). The poor result demonstrates that we can not leverage $f_{\bm{\phi}}$ as a binary classifier to capture the on-surface prior.

We further resolve this issue by learning $f_{\bm{\phi}}$ as unsigned distance functions. Using $l_i$ in the training set $T$ as continuous unsigned distances rather than discrete binary labels, we capture a more robust on-surface prior using $f_{\bm{\phi}}$ by minimizing the following loss function,\vspace{-0.1in}

\begin{equation}
\label{eq:2}
\begin{aligned}
\min_{\bm{\phi}} \frac{1}{I}\sum_{i\in[1,I]}||f_{\bm{\phi}}(\bm{p}_i,\bm{t}_i)-l_i||^2.
\end{aligned}
\end{equation}

We evaluate the on-surface prior learned with Eq.~(\ref{eq:2}) in Fig.~\ref{fig:Surfacepoint} (b). We leverage the same set of points used in Fig.~\ref{fig:Surfacepoint} (a) in the evaluation. We use a small unsigned distance threshold to filter out the points that are regarded as on-surface points by $f_{\bm{\phi}}$. The smooth surface shown by the on-surface points in Fig.~\ref{fig:Surfacepoint} (b) demonstrates that $f_{\bm{\phi}}$ is an eligible on-surface prior.

We obtain poor results by learning $f_{\bm{\phi}}$ as signed distance functions. This is because the sign information among different regions $\bm{t}_i$ is very complex, which makes it hard to learn the on-surface prior. We will compare this option in experiments later.

\noindent\textbf{Geometric Regularization. }One remaining question is that, is the on-surface prior adequate to learn SDFs $f_{\bm{\theta}}$ for unseen sparse point clouds $\bm{G}$ without ground truth signed distances and point normals? To justify this, we learn $f_{\bm{\theta}}$ by pushing all query projections to arrive on the surface according to the on surface evaluation below, where the learned on-surface prior is represented by the fixed parameters $\bm{\phi}$ in ODF $f_{\bm{\phi}}$,\vspace{-0.2in}

\begin{equation}
\label{eq:3}
\begin{aligned}
\min_{\bm{\theta}} \frac{1}{|Q|}\sum_{\bm{q}\in Q} |f_{\bm{\phi}}(\bm{q}-s\bm{d},Knn(\bm{q}-s\bm{d}))|,
\vspace{-0.2in}
\end{aligned}
\end{equation}

\noindent where $Q$ is a set of queries sampled around the sparse point cloud $\bm{G}$, $|Q|$ is the query number, $Knn(\bm{q})$ are $K$ nearest points on $\bm{G}$ of $\bm{q}$.

We reconstruct surfaces described by the learned $f_{\bm{\theta}}$ using the marching cubes algorithm. The poor surface in Fig.~\ref{fig:Surfacepoint} (d) demonstrates that $f_{\bm{\theta}}$ can not learn a correct signed distance field. The reason is that the first constraint defined in Eq.~(\ref{eq:2}) only constrains the query projections to be on the surface, while it does not care how the projection path provided by $f_{\bm{\theta}}$ should be. This results in an inaccurate or even wrong signed distance field.

We resolve this issue by introducing another constraint as a geometric regularization. The geometric regularization encourages the projection path to be the shortest, which matches the definition of signed distances, as defined below,\vspace{-0.1in}

\begin{equation}
\label{eq:4}
\begin{aligned}
\min_{\bm{\theta}} \frac{1}{|Q|}\sum_{\bm{q}\in Q} |f_{\bm{\theta}}(\bm{q})|.
\end{aligned}
\end{equation}

We visualize the effect of the geometric regularization in Fig.~\ref{fig:Surfacepoint} (e). Compared to the surface reconstruction without the geometric regularization in Fig.~\ref{fig:Surfacepoint} (d), the geometric regularization can infer a more accurate signed distance field, which leads to surface reconstruction with higher fidelity.

\noindent\textbf{Loss Function. }Our loss function pushes SDFs $f_{\bm{\theta}}$ to project queries $\bm{q}$ onto a surface along the shortest projection path. We learn $f_{\bm{\theta}}$ from a sparse point cloud $\bm{G}$ with the on-surface prior by combining Eq.~(\ref{eq:3}) and Eq.~(\ref{eq:4}) below, where $\lambda$ is a balance weight,\vspace{-0.15in}

\begin{equation}
\label{eq:5}
\begin{aligned}
\min_{\bm{\theta}} \frac{1}{|Q|}\sum_{\bm{q}\in Q}(| f_{\bm{\phi}}(\bm{q}-s\bm{d},Knn(\bm{q}-s\bm{d}))|+\lambda|f_{\bm{\theta}}(\bm{q})|).
\end{aligned}
\end{equation}

\noindent\textbf{Implementation. }We set $\lambda=0.4$ to balance the two constraints in Eq.~(\ref{eq:5}). We leverage the same network architectures as NP~\cite{Zhizhong2021icml} to learn the function $f_{\bm{\theta}}(\bm{q})$ and $f_{\bm{\phi}}(p,t)$. Additionally, we leverage an MLP with $8$ layers in $f_{\bm{\phi}}$ to learn the feature of $K$ nearest neighbors $\bm{t}=Knn(\bm{p})$ of $\bm{p}$. In addition, we regard the point $\bm{p}$ as the origin, and normalize the coordinates of $\bm{t}$ on the sparse point clouds according to the coordinates of $\bm{p}$, such that,\vspace{-0.1in}

\begin{equation}
\label{eq:1s}
\begin{aligned}
\bm{t}\gets\bm{t}-\bm{p} \quad \text{and} \quad \bm{p}\gets\bm{p}-\bm{p}.
\end{aligned}
\end{equation}

The purpose of this normalization is to make the on-surface prior learned from various regions on different shapes comparable. Note that we conduct this normalization in both learning $f_{\bm{\phi}}(\bm{p},\bm{t})$ in Eq.~(\ref{eq:2}) and leveraging the learned $f_{\bm{\phi}}$ as a prior in Eq.~(\ref{eq:3}) and Eq.~(\ref{eq:5}).

\section{Experiments}
\subsection{Setup}
\noindent\textbf{Dataset. }We evaluate our method in surface reconstruction for shapes and scenes. For shapes, we leverage a subset of ShapeNet~\cite{shapenet2015} with the same train and test splitting as~\cite{Zhizhong2021icml,liu2020meshing}. To evaluate our generalization ability, we employ our trained model to produce results under another unseen subset of ShapeNet~\cite{shapenet2015}. For scenes, we report our results under SceneNet~\cite{7780811Handa}, 3D Scene~\cite{DBLP:journals/tog/ZhouK13}, and Paris-rue-Madame~\cite{DBLP:conf/icpram/SernaMGD14}, where the latter two are real scanning datasets.

\noindent\textbf{Details. }In surface reconstruction for shapes, we uniformly sample $500$ points on each shape as sparse point clouds in both training and test sets. We leverage the training set of each class to form the dataset $T$. To learn $f_{\bm{\phi}}$ from $T$, we sample dense points on and around each shape as queries $\bm{q}_i$. Each query is paired with its $K=50$ nearest points in the sparse point cloud, and the $K=50$ nearest points form a local region $\bm{t}_i$. Moreover, we also calculate the unsigned distance $l_i$ for each $\bm{q}_i$. With the on-surface prior provided by the learned $f_{\bm{\phi}}$, we learn SDFs $f_{\bm{\theta}}$ for each single sparse point cloud in the test set by overfitting the shape without using the condition $\bm{c}$.

To reconstruct surfaces for scenes, we leverage the $f_{\bm{\phi}}$ learned from table class in ShapeNet as the on-surface prior to learn $f_{\bm{\theta}}$ for each single scene. To evaluate our performance under different point densities, we sample different numbers of points as the sparse point clouds. In 3D Scene dataset, we uniformly sample $100$, $500$, and $1000$ points per $m^2$ to form the sparse point clouds for each scene, while we uniformly sample $20$ and $100$ points per $m^2$ for each scene in SceneNet dataset. In Paris-rue-Madame, the point cloud containing $10M$ points was obtained by scanning on a street. We randomly sample $1M$ points as the sparse input.

\noindent\textbf{Metric. }We leverage L1 Chamfer Distance (L1CD), L2 Chamfer Distance (L2CD), Normal Consistency (NC), and F-Score with a threshold of $0.001$ for shapes and $0.025$ for scenes. For CD, we sample $100K$ points on both the reconstructed and the ground truth surfaces for single shapes under ShapeNet, while sampling $1M$ points for scenes.

\subsection{Surface Reconstruction on Shapes}
\noindent\textbf{ShapeNet. }We first evaluate our method under ShapeNet. We leverage the pretrained models of COcc~\cite{Peng2020ECCV}, LIG~\cite{jiang2020lig}, and ISO~\cite{yifan2021iso} to produce their results on shapes with $500$ points, where we also provide COcc and LIG the ground truth point normals. We tried to retrain these methods using the same sparse point clouds as ours, but we failed to produce better results. We produce the results of NP~\cite{Zhizhong2021icml} by retraining it with the same shapes with $500$ points as ours.

\begin{figure}[tb]
  \centering
   \includegraphics[width=\linewidth]{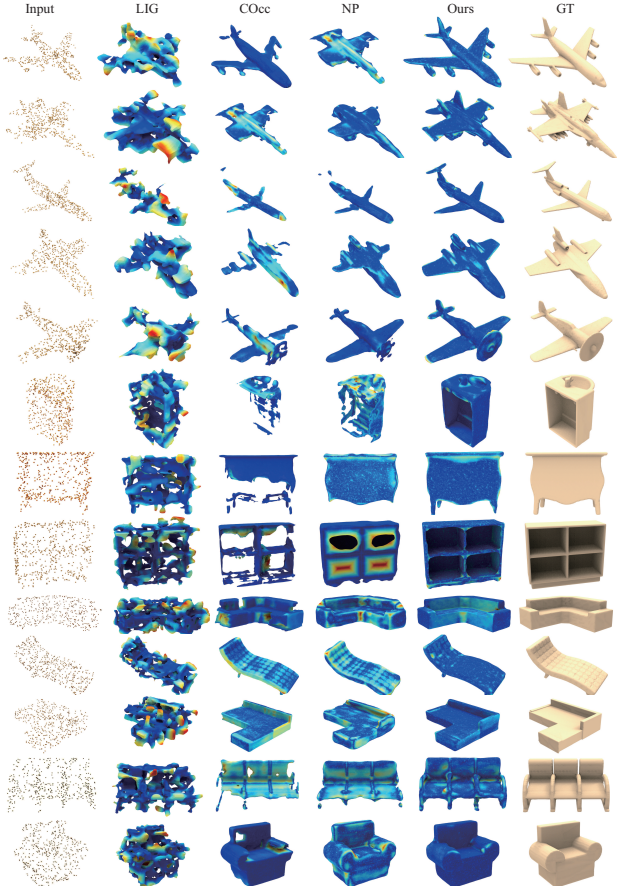}
  %
  %
  \vspace{-0.3in}
\caption{\label{fig:shapenet}Visual comparison with LIG~\cite{jiang2020lig}, COcc~\cite{Peng2020ECCV}, NP~\cite{Zhizhong2021icml} under ShapeNet. The input contains $500$ points. Color indicates point to surface errors.}
\vspace{-0.1in}
\end{figure}

We report the numerical comparison under $8$ classes in Tab.~\ref{table:l1shapenet},~\ref{table:ncshapenet}, and~\ref{table:F-Score}. We achieve the best results in terms of all the three metrics in all shape classes. We found that current state-of-the-art methods are still struggling to reveal surfaces from sparse point clouds, while our method can handle the sparseness of point clouds well with the on-surface prior. We further demonstrate our advantages using surface reconstruction with error maps in Fig.~\ref{fig:shapenet}. The visual comparison indicates that the latest methods are dramatically affected by the sparseness of points, which results in poor and incomplete surfaces. While our method is able to reveal surfaces from sparse point clouds in higher accuracy.

\begin{table}[t]
\centering
\resizebox{\linewidth}{!}{
    \begin{tabular}{c|ccccc}  
     \hline
     Class&NP~\cite{Zhizhong2021icml}&COcc~\cite{Peng2020ECCV}&LIG~\cite{jiang2020lig}&ISO~\cite{yifan2021iso}&\textbf{Ours}\\
     \hline
     airplane&0.082&0.078&0.185&0.776&\textbf{0.076}\\
     cabinet&0.101&0.133&0.188&0.783&\textbf{0.068}\\
     chair&0.163&0.167&0.183&0.822&\textbf{0.083}\\
     display&0.074&0.118&0.183&0.727&\textbf{0.068}\\
     lamp&0.087&0.187&0.207&1.146&\textbf{0.066}\\
     sofa&0.088&0.101&0.178&0.649&\textbf{0.071}\\
     table&0.187&0.187&0.187&0.827&\textbf{0.053}\\
     vessele&0.063&0.089&0.179&0.661&\textbf{0.057}\\
     \hline
     mean&0.106&0.132&0.186&0.799&\textbf{0.068}\\
     \hline
\end{tabular}
   }
   \vspace{-0.15in}
   \caption{L1CD$\times 10$ comparison under ShapeNet.}
   \vspace{-0.25in}
   \label{table:l1shapenet}
\end{table}

\begin{table}[t]
\centering
\resizebox{\linewidth}{!}{
    \begin{tabular}{c|ccccc}  
     \hline
     Class&NP~\cite{Zhizhong2021icml}&COcc~\cite{Peng2020ECCV}&LIG~\cite{jiang2020lig}&ISO~\cite{yifan2021iso}&\textbf{Ours}\\
     \hline
     airplane&0.863&0.850&0.722&0.638&\textbf{0.897}\\
     cabinet&0.850&0.816&0.657&0.545&\textbf{0.888}\\
     chair&0.840&0.829&0.707&0.610&\textbf{0.864}\\
     display&0.901&0.885&0.677&0.566&\textbf{0.930}\\
     lamp&0.885&0.844&0.733&0.689&\textbf{0.892}\\
     sofa&0.878&0.844&0.681&0.589&\textbf{0.905}\\
     table&0.806&0.845&0.684&0.578&\textbf{0.907}\\
     vessele&0.839&0.799&0.682&0.623&\textbf{0.866}\\
     \hline
     mean&0.858&0.839&0.693&0.605&\textbf{0.894}\\
     \hline
\end{tabular}
   }
   \vspace{-0.15in}
   \caption{NC comparison under ShapeNet.}
   \vspace{-0.15in}
   \label{table:ncshapenet}
\end{table}

\begin{table}[t]
\centering
\resizebox{\linewidth}{!}{
    \begin{tabular}{c|ccccc}  
     \hline
     Class&NP~\cite{Zhizhong2021icml}&COcc~\cite{Peng2020ECCV}&LIG~\cite{jiang2020lig}&ISO~\cite{yifan2021iso}&\textbf{Ours}\\
     \hline
     airplane&0.977&0.971&0.759&0.422&\textbf{0.989}\\
     cabinet&0.955&0.891&0.793&0.387&\textbf{0.983}\\
     chair&0.914&0.886&0.791&0.354&\textbf{0.962}\\
     display&0.946&0.936&0.793&0.443&\textbf{0.959}\\
     lamp&0.971&0.897&0.739&0.339&\textbf{0.975}\\
     sofa&0.911&0.918&0.815&0.475&\textbf{0.926}\\
     table&0.823&0.816&0.788&0.369&\textbf{0.836}\\
     vessele&0.988&0.963&0.797&0.537&\textbf{0.989}\\
     \hline
     mean&0.936&0.909&0.784&0.416&\textbf{0.952}\\
     \hline
\end{tabular}
   }
   \vspace{-0.15in}
   \caption{F-Score comparison under ShapeNet.}
   \vspace{-0.25in}
   \label{table:F-Score}
\end{table}

\begin{figure}[tb]
  \centering
   \includegraphics[width=\linewidth]{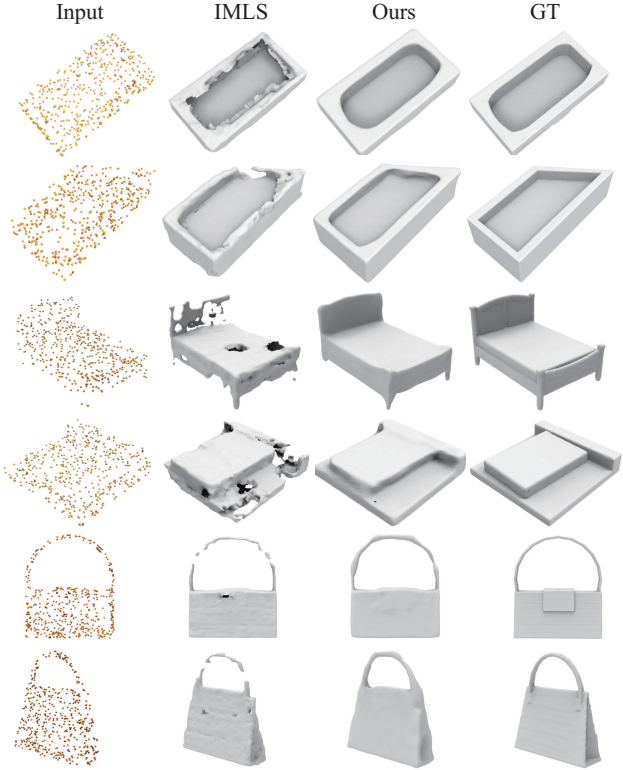}
  %
  %
  \vspace{-0.3in}
\caption{\label{fig:shapenetunseen}Visual comparison with IMLS~\cite{Liu2021MLS} in unseen classes under ShapeNet.}
\vspace{-0.3in}
\end{figure}

To evaluate the generalization ability of our learned on-surface prior, we leverage the $f_{\bm{\phi}}$ learned under table class to reconstruct surfaces for sparse point clouds from $5$ unseen classes in Tab.~\ref{table:generalization}. The numerical comparison with IMLS~\cite{Liu2021MLS} shows that we can leverage the learned class-agnostic and object-agnostic on-surface prior to reconstruct more accurate surfaces for unseen point clouds, which demonstrates our better generalization ability than IMLS. Note that we leverage the pretrained model of IMLS that was trained under $13$ shape classes and evaluate it under the same $5$ unseen classes with the same input as ours. We further highlight our advantages in visual comparison with IMLS in Fig.~\ref{fig:shapenetunseen}.

\subsection{Surface Reconstruction on Scenes}
We leverage the $f_{\bm{\phi}}$ learned from table class in ShapeNet as the on-surface prior for scenes. We report our results by learning $f_{\bm{\theta}}$ to overfit each scene.

\begin{figure*}[tb]
  \centering
   \includegraphics[width=\linewidth]{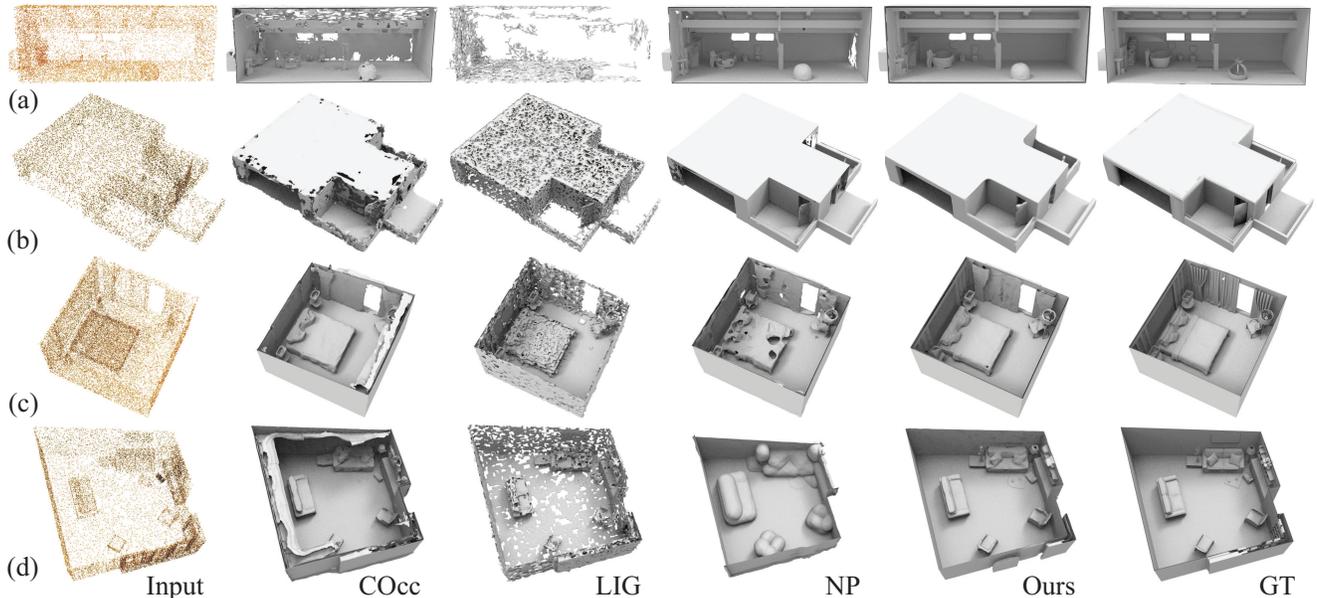}
  %
  %
  \vspace{-0.3in}
\caption{\label{fig:SceneNet}Visual comparison with COcc~\cite{Peng2020ECCV}, LIG~\cite{jiang2020lig}, NP~\cite{Zhizhong2021icml} under SceneNet. The input contains $20$ points/$m^2$ in (a) and (b), and $100$ points/$m^2$ in (c) and (d). More comparisons can be found in our supplemental materials.}
\vspace{-0.15in}
\end{figure*}

\noindent\textbf{SceneNet. }We further evaluate our surface reconstruction performance under SceneNet~\cite{7780811Handa}. With different point densities, we evaluate our surface reconstruction accuracy in different metrics. We leverage the pretrained models of COcc~\cite{Peng2020ECCV} and LIG~\cite{jiang2020lig} to produce their results and retrain NP~\cite{Zhizhong2021icml} using the same input point clouds as ours. The numerical comparison in Tab.~\ref{table:t12scenenet} demonstrates that our method significantly outperforms other methods even there are only $20$ points per /$m^2$ in each scene. We further demonstrate our advantages over the state-of-the-art in Fig.~\ref{fig:SceneNet}. Visual comparison indicates that current methods cannot produce smooth and complete surfaces from sparse point clouds, while we show our superior performance over them by producing surfaces with more detailed geometry.

\begin{figure*}[tb]
  \centering
   \includegraphics[width=\linewidth]{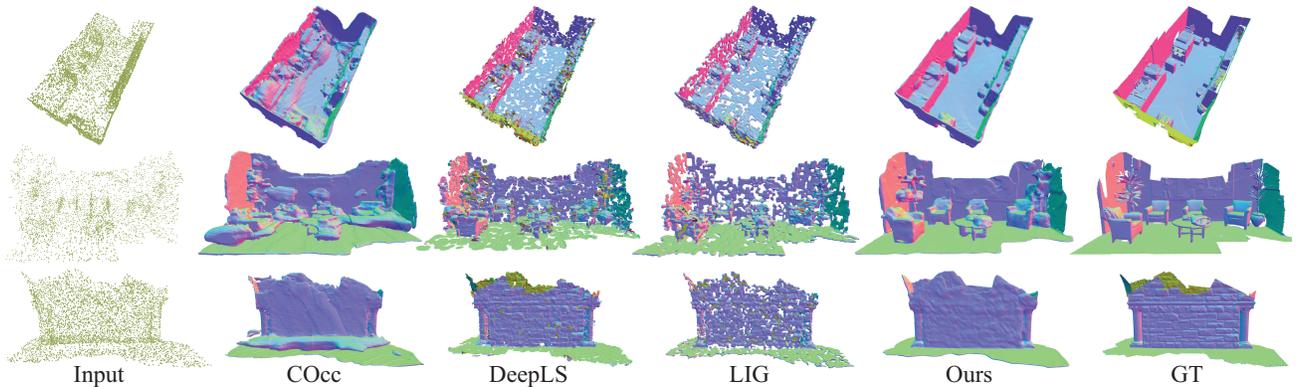}
  %
  %
  \vspace{-0.3in}
\caption{\label{fig:3DScanNetEPS}Visual comparison with COcc~\cite{Peng2020ECCV}, DeepLS~\cite{DBLP:conf/eccv/ChabraLISSLN20}, LIG~\cite{jiang2020lig} under 3D Scene dataset. The input contains $100$ points/$m^2$. Color indicates normals.}
\vspace{-0.22in}
\end{figure*}

\begin{table*}[tb]
\centering
\resizebox{\linewidth}{!}{
    \begin{tabular}{c|c|c|c||c|c|c||c|c|c||c|c|c||c|c|c}
     \hline

        &\multicolumn{3}{c||}{bed}&\multicolumn{3}{c||}{bag}&\multicolumn{3}{c||}{bathtub}&\multicolumn{3}{c||}{bottle}&\multicolumn{3}{c}{pillow}\\
        \hline
        &L1CD&NC&FScore&L1CD&NC&FScore&L1CD&NC&FScore&L1CD&NC&FScore&L1CD&NC&FScore\\
     \hline
     IMLS~\cite{Liu2021MLS}&0.077&0.838&0.963&0.058&0.926&0.981&0.069&0.889 &0.977 &0.044 &0.954 &0.994 &\textbf{0.039} &0.959 &0.997\\
     Ours&\textbf{0.069}&\textbf{0.908}&\textbf{0.981}&\textbf{0.054}&\textbf{0.936}&\textbf{0.990}&\textbf{0.056}&\textbf{0.952}&\textbf{0.993}&\textbf{0.043}&\textbf{0.977}&\textbf{0.995}&\textbf{0.039}&\textbf{0.970}&\textbf{0.999}\\
     \hline
   \end{tabular}}
   \vspace{-0.15in}
   \caption{Generalization ability evaluation under ShapeNet. L1CD$\times 10$.}
   \vspace{-0.12in}
   \label{table:generalization}
\end{table*}

\begin{table*}[tb]
\centering
\resizebox{\linewidth}{!}{
    \begin{tabular}{c|c|c|c|c||c|c|c||c|c|c||c|c|c||c|c|c||c|c|c}
     \hline

        &&\multicolumn{3}{c||}{Livingroom}&\multicolumn{3}{c||}{Bathroom}&\multicolumn{3}{c||}{Bedroom}&\multicolumn{3}{c||}{Kitchen}&\multicolumn{3}{c||}{Office}&\multicolumn{3}{c}{Mean}\\
        \hline
        &&L1CD&NC&FScore&L1CD&NC&FScore&L1CD&NC&FScore&L1CD&NC&FScore&L1CD&NC&FScore&L1CD&NC&FScore\\
        \hline
        \multirow{4}{*}{\rotatebox{90}{20/$m^2$}}&LIG~\cite{jiang2020lig}&0.032&0.719&0.790&0.030&0.737&0.807&0.029&0.735&0.818&0.029&0.727&0.817&0.033&0.737&0.805&0.030&0.730&0.808\\
        &COcc~\cite{Peng2020ECCV}&0.026&0.895&0.955&0.025&0.862&0.988&0.028&0.823&0.976&0.028&0.849&0.982&0.030&0.829&0.958&0.027&0.852&0.971\\
        &NP~\cite{Zhizhong2021icml}&0.068&0.827&0.718&0.072&0.716&0.658&0.044&0.782&0.740&0.069&0.720&0.689&0.066&0.834&0.663&0.037&0.776&0.693\\
        \cline{2-20}
        &Ours&\textbf{0.025}&\textbf{0.904}&\textbf{0.961}&\textbf{0.018}&\textbf{0.924}&\textbf{0.991}&\textbf{0.023}&\textbf{0.919}&\textbf{0.976}&\textbf{0.025}&\textbf{0.911}&\textbf{0.983}&\textbf{0.029}&\textbf{0.851}&\textbf{0.967}&\textbf{0.024}&\textbf{0.902}&\textbf{0.975}\\
     \hline
        \multirow{4}{*}{\rotatebox{90}{100/$m^2$}}&LIG~\cite{jiang2020lig}&0.019&0.922&0.919&0.018&0.930&0.915&0.017&0.918&0.920&0.016&0.920&0.936&0.020&0.910&0.936&0.018&0.920&0.925\\
        &COcc~\cite{Peng2020ECCV}&0.026&0.895&0.979&0.025&0.910&0.979&0.026&0.890&0.980&0.027&0.898&0.981&0.027&0.894&0.985&0.026&0.897&0.981\\
        &NP~\cite{Zhizhong2021icml}&0.069&0.883&0.799&0.028&0.907&0.893&0.032&0.890&0.878&0.042&0.896&0.838&0.066&0.866&0.733&0.047&0.888&0.828\\
        \cline{2-20}
        &Ours&\textbf{0.018}&\textbf{0.960}&\textbf{0.985}&\textbf{0.015}&\textbf{0.947}&\textbf{0.984}&\textbf{0.013}&\textbf{0.960}&\textbf{0.983}&\textbf{0.012}&\textbf{0.950}&\textbf{0.985}&\textbf{0.019}&\textbf{0.921}&\textbf{0.990}&\textbf{0.015}&\textbf{0.947}&\textbf{0.985}\\
     \hline
   \end{tabular}}
   \vspace{-0.15in}
   \caption{Numerical comparison in surface reconstruction under SceneNet.}
   \vspace{-0.12in}
   \label{table:t12scenenet}
\end{table*}

\noindent\textbf{3D Scene. }We leverage the same strategy to evaluate our method under 3D scenes dataset. With different numbers of points in the input point clouds, we compare our method with COcc~\cite{Peng2020ECCV}, LIG~\cite{jiang2020lig} and DeepLS~\cite{DBLP:conf/eccv/ChabraLISSLN20}. With the ground truth point normals, we report the results of COcc and LIG using their pretrained models, and we retrain DeepLS by overfitting each point cloud. The numerical comparison in Tab.~\ref{table:t12100} demonstrates our superior performance in all scene classes with different point densities. We further highlight our performance in the visual comparison in Fig.~\ref{fig:3DScanNetEPS}. The visual comparison with $100$ points/$m^2$ indicates that the latest methods still struggle to reveal surfaces from sparse point clouds while ours can produce more accurate surfaces.

\begin{figure*}[tb]
  \centering
   \includegraphics[width=\linewidth]{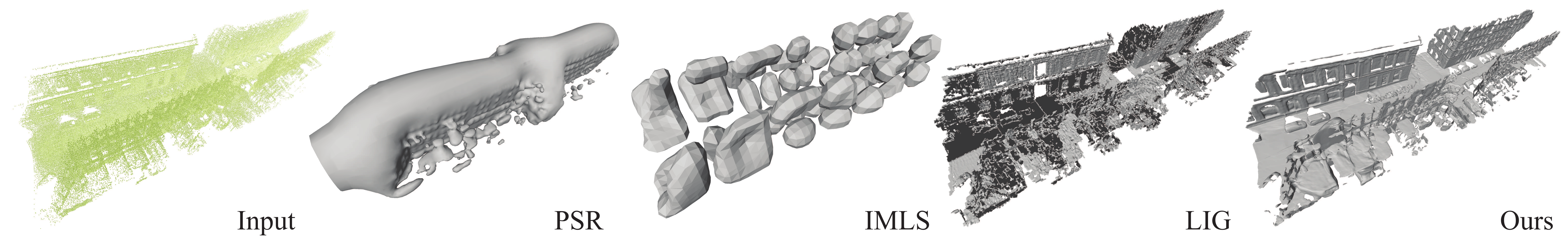}
  %
  %
  \vspace{-0.3in}
\caption{\label{fig:ParisStreet}Visual comparison with PSR~\cite{journals/tog/KazhdanH13}, LIG~\cite{jiang2020lig}, and IMLS~\cite{Liu2021MLS} under real scanning.}
\vspace{-0.25in}
\end{figure*}

\noindent\textbf{Paris-rue-Madame. }Finally, we evaluate our method using a large-scale real scanning. We split the point cloud into multiple sections and leverage each part to train our method or evaluate other methods. We visualize our reconstruction of the entire scene and some partial sections in Fig.~\ref{fig:ParisStreet} and Fig.~\ref{fig:ParisStreetpart}, respectively. The visual comparison with the state-of-the-art demonstrates that our method can reconstruct more accurate surfaces from sparse point clouds.

These plausible results under scenes indicate that our method can reveal surfaces for scenes with complex geometry details, and our on-surface prior has remarkable generalization ability.

\begin{table*}[!]
\centering
\resizebox{\linewidth}{!}{
    \begin{tabular}{c|c|c|c|c||c|c|c||c|c|c||c|c|c||c|c|c}
     \hline

        &&\multicolumn{3}{c||}{Burghers}&\multicolumn{3}{c||}{Lounge}&\multicolumn{3}{c||}{Copyroom}&\multicolumn{3}{c||}{Stonewall}&\multicolumn{3}{c}{Totempole}\\
        \hline
        &&L2CD&L1CD&NC&L2CD&L1CD&NC&L2CD&L1CD&NC&L2CD&L1CD&NC&L2CD&L1CD&NC\\
     \hline
     \multirow{4}{*}{\rotatebox{90}{100/$m^2$}}&COcc~\cite{Peng2020ECCV}&8.904&0.040&0.890&6.979 &0.041 &0.884 &6.78 &0.041 &0.856 &12.22 &0.051 &0.903 &4.412 &0.041 &0.874\\
     &LIG~\cite{jiang2020lig}&3.112 &0.044 &0.839 &9.128 &0.054 &0.833 &4.363 &0.039 &0.804&5.143 &0.046 &0.853 &9.58 &0.062 &0.887\\   &DeepLS~\cite{DBLP:conf/eccv/ChabraLISSLN20}&3.111&0.050&0.856&3.894&0.056&0.764&1.498&0.033&0.777&2.427&0.038&0.885&4.214&0.043&\textbf{0.908}\\     \cline{2-17}
     &Ours&\textbf{0.544}&\textbf{0.018}&\textbf{0.922}&\textbf{0.435}&\textbf{0.013}&\textbf{0.929}&\textbf{0.434}&\textbf{0.017}&\textbf{0.911}&\textbf{0.371}&\textbf{0.016}&\textbf{0.950}&\textbf{3.986}&\textbf{0.040}&0.889\\
     \hline
     \hline
     \multirow{4}{*}{\rotatebox{90}{500/$m^2$}}&COcc~\cite{Peng2020ECCV}&26.97&0.081&0.905&9.044 &0.046 &0.894 &10.08 &0.046 &0.885 &17.70 &0.063 &0.909 &2.165 &\textbf{0.024} &0.937\\
     &LIG~\cite{jiang2020lig}&3.080 &0.046 &0.840 &6.729 &0.052 &0.831 &4.058 &0.038 &0.810&4.919 &0.043 &0.878 &9.38 &0.062 &0.887\\
     &DeepLS~\cite{DBLP:conf/eccv/ChabraLISSLN20}&0.714&0.020&0.923&10.88&0.077&0.814&0.552&0.015&0.907&0.673&0.018&0.951&21.15&0.122&0.927\\
     \cline{2-17}
     &Ours&\textbf{0.609}&\textbf{0.018}&\textbf{0.930}&\textbf{0.529}&\textbf{0.013}&\textbf{0.926}&\textbf{0.483}&\textbf{0.014}&\textbf{0.908}&\textbf{0.666}&\textbf{0.013}&\textbf{0.955}&\textbf{2.025}&0.041&\textbf{0.954}\\
     \hline
     \hline
     \multirow{4}{*}{\rotatebox{90}{1000/$m^2$}}&COcc~\cite{Peng2020ECCV}&27.46&0.079&0.907&9.54 &0.046 &0.894 &10.97 &0.045 &0.892 &20.46 &0.069 &0.905 &2.054 &0.021 &0.943\\
     &LIG~\cite{jiang2020lig}&3.055 &0.045 &0.835 &9.672 &0.056 &0.833 &3.61 &0.036 &0.810&5.032 &0.042 &0.879 &9.58 &0.062 &0.887\\
    &DeepLS~\cite{DBLP:conf/eccv/ChabraLISSLN20}&\textbf{0.401}&\textbf{0.017}&0.920&6.103&0.053&0.848&0.609&0.021&0.901&0.320&0.015&0.954&\textbf{0.601}&\textbf{0.017}&0.950\\
     \cline{2-17}
     &Ours&1.339&0.031&\textbf{0.929}&\textbf{0.432}&\textbf{0.014}&\textbf{0.934}&\textbf{0.405}&\textbf{0.014}&\textbf{0.914}&\textbf{0.266}&\textbf{0.014}&\textbf{0.957}&1.089&0.029&\textbf{0.954}\\
   \hline
   \end{tabular}}
   \vspace{-0.15in}
   \caption{Surface reconstruction for point clouds under 3D Scene. L2CD$\times 1000$.}
   \vspace{-0.180in}
   \label{table:t12100}
\end{table*}

\subsection{Ablation Study}
We conduct ablation studies to justify the effectiveness of our method. We report numerical comparison in surface reconstruction, using $10$ shapes to learn $f_{\bm{\phi}}$ during training and using another $10$ shapes to learn $f_{\bm{\theta}}$ during testing, both of which are from ABC dataset~\cite{Koch_2019_CVPR}. We sample $500$ points on each shape.

\noindent\textbf{Prior. }We first justify the effectiveness of our on-surface prior in Tab.~\ref{table:NOX11prior}. We learn a signed distance function (``SDF'') as the on-surface prior, and directly leverage it to reconstruct surfaces without learning $f_{\bm{\theta}}$. We found that SDF can not reconstruct plausible surfaces, even with $f_{\bm{\theta}}$ (``SDF+$f_{\bm{\theta}}$''). The reason is that the on-surface prior is learned from various small regions without normalizing orientation. This makes the network hard to determine the sign of the distances. We can also learn the prior as a binary classifier, but the result (``Binary+$f_{\bm{\theta}}$'') gets worse. Our method learns on-surface prior as an unsigned distance function (``UDF''), which achieves the best performance.

\begin{table}[tb]
\centering
\resizebox{\linewidth}{!}{
    \begin{tabular}{c|c|c|c|c}  
     \hline
          &SDF&SDF+$f_{\bm{\theta}}$&Binary+$f_{\bm{\theta}}$&UDF+$f_{\bm{\theta}}$\\   
     \hline
       L1CD&0.050&0.049&0.083&\textbf{0.015}\\ 
       NC&0.827&0.833&0.743&\textbf{0.928}\\
       \hline
   \end{tabular}}
   \vspace{-0.15in}
   \caption{Effect of on-surface prior.}
   \vspace{-0.15in}
   \label{table:NOX11prior}
\end{table}

\noindent\textbf{Balance Weights. }Then, we explore the effect of balance weight $\lambda$ in Eq.~(\ref{eq:5}). We try different candidates including $\{0,0.2,0.4,0.6\}$, and report the results in Tab.~\ref{table:NOX11balance}. The results of ``0'' are the worst, which highlights the importance of the geometric regularization in Eq.~(\ref{eq:4}). If the weight is too large (``0.6''), it turns to encourage $f_{\bm{\theta}}$ to output $0$ for any query locations. If the weight is too small (``0.2'') , the geometry regularization can not constrain $f_{\bm{\theta}}$ to predict minimum distances to the surface. We found ``0.4'' is a proper tradeoff.

\begin{table}[tb]
\centering
    \begin{tabular}{c|c|c|c|c}  
     \hline
          $\lambda$& 0&0.2&0.4&0.6\\   
     \hline
       L1CD&0.050&0.016 &\textbf{0.015}&0.023\\ 
       NC&0.569&0.894 &\textbf{0.928}&0.890\\
       \hline
   \end{tabular}
   \vspace{-0.15in}
   \caption{Effect of balance weight $\lambda$.}
   \vspace{-0.28in}
   \label{table:NOX11balance}
\end{table}

\noindent\textbf{$K$ Nearest Neighbors. }We explore the effect of $K$ on the on-surface prior learned by $f_{\bm{\phi}}$. We try different $K$ to form the nearest region on sparse point clouds for each query, such as $\{25,50,100,200\}$. The numerical comparison in Tab.~\ref{table:NOX11knn} indicates that $K$ slightly affects the performance, and achieves the best with $50$ neighboring points.

\begin{table}[tb]
\centering
    \begin{tabular}{c|c|c|c|c}  
     \hline
          $K$& 25&50&100&200\\   
     \hline
       L1CD&0.016&\textbf{0.015}&0.016&0.018\\ 
       NC&0.745&\textbf{0.928}&0.919&0.897\\
       \hline
   \end{tabular}
   \vspace{-0.15in}
   \caption{Effect of $K$ nearest neighbors.}
   \vspace{-0.15in}
   \label{table:NOX11knn}
\end{table}

\noindent\textbf{Normalization. }We explore different strategies to normalize the input of $f_{\bm{\phi}}$, i.e., $\bm{p}$ and its $K$ nearest neighbors $\bm{t}$, by translating or rotating when learning $f_{\bm{\phi}}$. We leverage Eq.~(\ref{eq:1s}) for translation. For rotation, we align the vector connecting $\bm{p}$ and its nearest point on $\bm{t}$ with the positive direction of axis Z. We report the results in Tab.~\ref{table:NOX11norm}. We found that the translation is the key to learning good prior as $f_{\bm{\phi}}$, while rotation does not help.

\begin{table}[tb]
\centering
\resizebox{\linewidth}{!}{
    \begin{tabular}{c|c|c|c|c}  
     \hline
          Normalization& No Trans, No Rot&Trans, No Rot&No Trans, Rot&Trans, Rot\\   
     \hline
       L1CD&0.084&\textbf{0.015}&0.086&0.035\\ 
       NC&0.766&\textbf{0.928}&0.745&0.843\\
       \hline
   \end{tabular}}
   \vspace{-0.1in}
   \caption{Effect of translation and rotation.}
   \vspace{-0.1in}
   \label{table:NOX11norm}
\end{table}

\noindent\textbf{Network. }We found that the MLP we leveraged to learn features of $K$ nearest neighbors $\bm{t}$ in $f_{\bm{\phi}}$ performs much better than PointNet~\cite{cvprpoint2017} and PointNet++~\cite{nipspoint17}, as demonstrated in Tab.~\ref{table:NOX11net}. We found PointNet and PointNet++ can not understand sparse points well due to the maxpooling while MLP takes full advantage of the point order sorted by the distances to achieve a remarkable performance.

\begin{table}[tb]
\centering
    \begin{tabular}{c|c|c|c}  
     \hline
          & PointNet~\cite{cvprpoint2017}&PointNet++~\cite{nipspoint17}&MLP\\   
     \hline
       L1CD&0.076&0.070&\textbf{0.015}\\ 
       NC&0.784&0.797&\textbf{0.928}\\
       \hline
   \end{tabular}
   \vspace{-0.1in}
   \caption{Effect of network.}
   \vspace{-0.1in}
   \label{table:NOX11net}
\end{table}

\begin{figure}[tb]
  \centering
   \includegraphics[width=\linewidth]{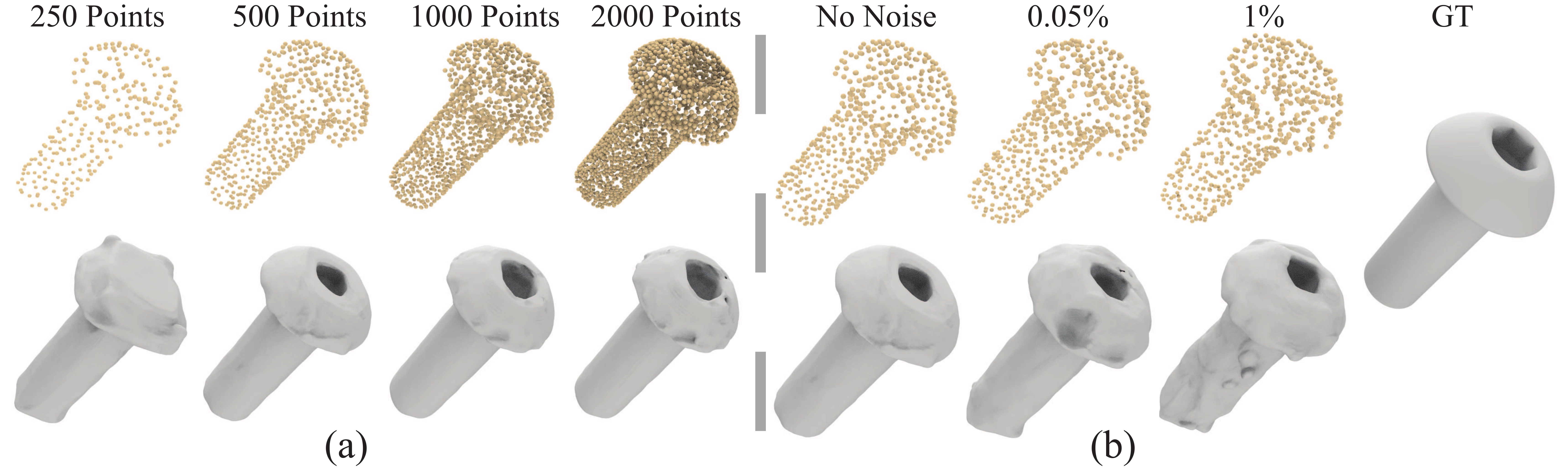}
  %
  %
  \vspace{-0.2in}
\caption{\label{fig:AblationStudies}Generalization ability of $f_{\bm{\phi}}$ to (a) point densities and (b) noisy points.}
\vspace{-0.2in}
\end{figure}

\noindent\textbf{Generalization. }We further evaluate the generalization ability of our on-surface prior learned by $f_{\bm{\phi}}$ to different point densities during test. We leverage the $f_{\bm{\phi}}$ trained with $K=50$ nearest neighbors on shapes represented by $500$ points to reconstruct surfaces for shapes with different point densities, such as $\{250,500,1000,2000\}$. Tab.~\ref{table:NOX11den} and Fig.~\ref{fig:AblationStudies} (a) demonstrate that $f_{\bm{\phi}}$ generalizes better to higher point densities than lower ones.

\begin{table}[tb]
\centering
    \begin{tabular}{c|c|c|c|c}  
     \hline
          Density&250&500&1000&2000\\   
     \hline
       L1CD&0.020&0.015&0.013&\textbf{0.012}\\ 
       NC&0.901&0.928&0.937&\textbf{0.939}\\
       \hline
   \end{tabular}
   \vspace{-0.1in}
   \caption{Generalization ability of $f_{\bm{\phi}}$ to point densities.}
   \vspace{-0.25in}
   \label{table:NOX11den}
\end{table}

\begin{figure}[tb]
  \centering
   \includegraphics[width=\linewidth]{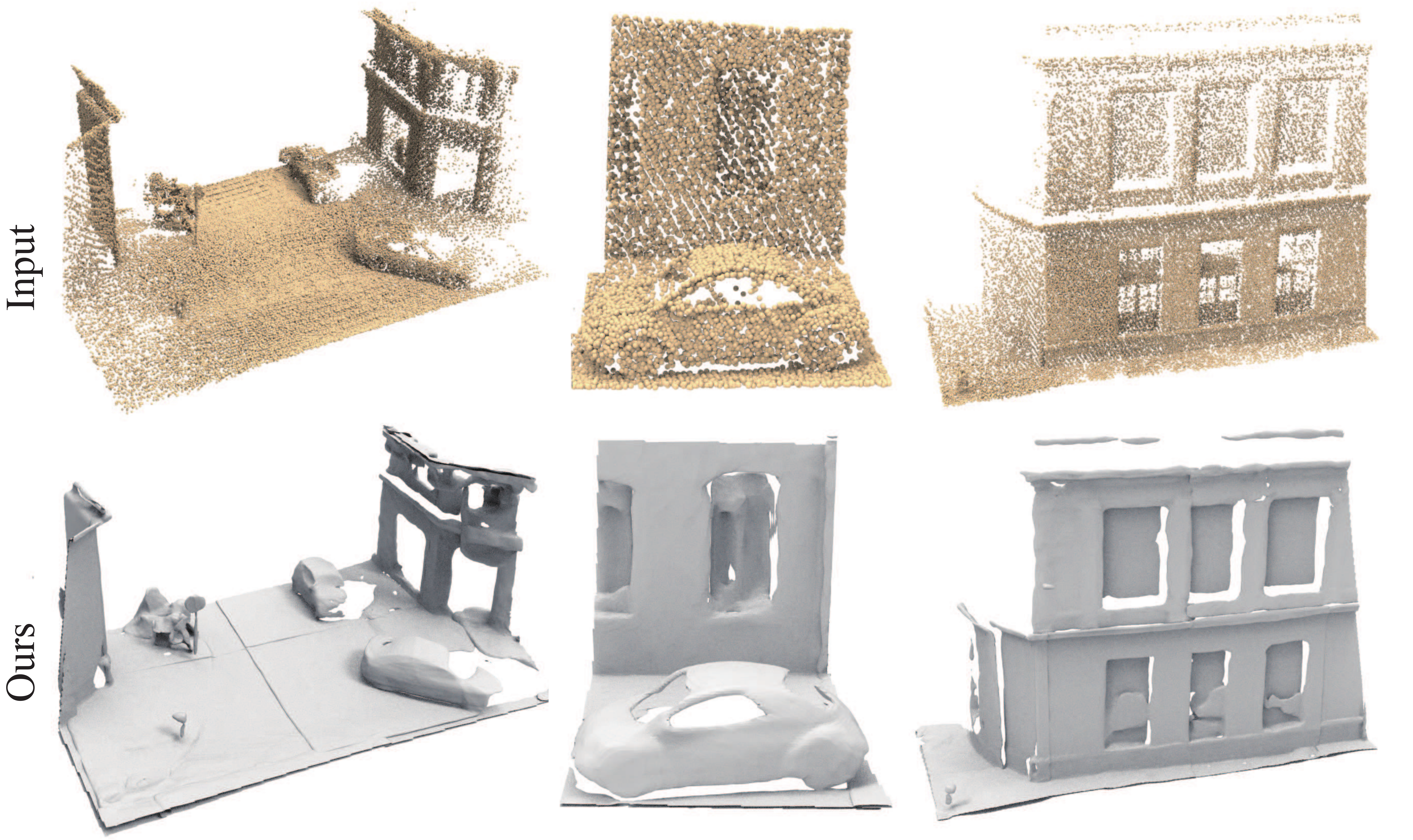}
  %
  %
  \vspace{-0.2in}
\caption{\label{fig:ParisStreetpart}Surface reconstruction from real scanning.}
\vspace{-0.1in}
\end{figure}

\noindent\textbf{Noise. }Besides our surface reconstruction under real scanning with noise in Fig.~\ref{fig:ParisStreetpart} and Fig.~\ref{fig:ParisStreet}, we further report the generalization ability of $f_{\bm{\phi}}$ to noise. We leverage the learned $f_{\bm{\phi}}$ to reconstruct surfaces from noisy point clouds with two standard deviations including $0.5\%$ and $1\%$. The comparison in Tab.~\ref{table:NOX11noisy} and Fig.~\ref{fig:AblationStudies} (b) demonstrates that $f_{\bm{\phi}}$ is able to generalize to different level noise.

\begin{table}[tb]
\centering
    \begin{tabular}{c|c|c|c}  
     \hline
          Noise&0\%&0.5\%&1\%\\   
     \hline
       L1CD&\textbf{0.015}&0.017&0.018\\ 
       NC&\textbf{0.928}&0.912&0.908\\
       \hline
   \end{tabular}
   \vspace{-0.1in}
   \caption{Generalization ability of $f_{\bm{\phi}}$ to noisy points.}
   \vspace{-0.1in}
   \label{table:NOX11noisy}
\end{table}

\noindent\textbf{Limitation. }Although we show our superior performance on sparse point clouds, we can not handle the incomplete point clouds, which is an extreme sparse senecio. It would be a good direction to combine the shape completion prior in the future work.

\section{Conclusion}
We resolve the issue of reconstructing surfaces from sparse point clouds, which surfers the state-of-the-art methods a lot. We achieve this by learning SDFs via overfitting a sparse point cloud with an on-surface prior. We successfully learn class-agnostic and object-agnostic on-surface prior to reveal surfaces from sparse point clouds in a data-driven way. Our method is able to further leverage the learned on-surface prior with a geometric regularization to learn highly accurate SDFs for unseen sparse point clouds. Our method does not require signed distances or point normals to learn SDFs and the learned on-surface prior demonstrates remarkable generalization ability. Our method outperforms the latest methods in surface reconstruction from sparse point clouds under different benchmarks.

{\small
\bibliographystyle{ieee_fullname}
\bibliography{PaperForReview}
}

\end{document}